\documentclass[sigconf]{acmart}

\renewcommand\footnotetextcopyrightpermission[1]{} 
\usepackage{booktabs}
\usepackage{filecontents}

\usepackage{graphicx}
\usepackage[many]{tcolorbox}

\usepackage{mathtools,amssymb,latexsym,amsfonts,stmaryrd}
\usepackage{pbox}
\usepackage{xfrac}
\usepackage{booktabs}
\usepackage{xcolor}
\usepackage{threeparttable}
\usepackage{amsthm}
\usepackage{hyperref}
\usepackage{multirow}
\usepackage{tikz}
\usepackage{mathrsfs}
\usepackage{amsmath}
\usepackage{mathpartir}
\usepackage{algorithm}
\usepackage{url}
\usepackage{syntax}
\usepackage{framed}
\usepackage[noend]{algpseudocode}
\usepackage{flushend}
\usepackage{listings}
\usepackage{moresize}
\usepackage{wrapfig}
 
\usepackage{seqsplit}
\usepackage{alltt}
\usepackage{xspace}
\usepackage{pgfplots}
\usepackage{subcaption}
\usetikzlibrary{positioning}
\usepackage{enumitem}
\usepackage{pifont}

\DeclareMathAlphabet{\mathcal}{OMS}{cmsy}{m}{n}

\usepackage{amsmath, listings, amsthm, amssymb, proof, xspace}

\usepackage{thmtools}

\declaretheoremstyle[spaceabove=\topsep,notefont=\normalfont\itshape]{mystyle}

\newcommand{\revise}[2]{{\color{red}{\ifx&#1&\else- #1\fi}} {\color{ForestGreen}{\ifx&#2&\else+ #2\fi}}}%
\renewcommand{\revise}[2]{#2}%

\usetikzlibrary{arrows,patterns, decorations.pathreplacing}

\newcommand{\F}{Fig.}
\newcommand{\T}{Table}
\renewcommand{\S}{Sec.}

\usepackage{makecell}

\newcommand{\ignore}[1]{}

\lstdefinestyle{base}{
  moredelim=**[is][\color{red}]{@}{@},
  escapeinside={<@}{@>}
}

\usepackage{amssymb}
\usepackage{pifont}

\newcommand\DejaVuttfamily{%
  \fontfamily{DejaVuSansMono-TLF}\selectfont }

\lstdefinestyle{base}{
  moredelim=**[is][\color{red}]{@}{@},
  escapeinside={<@}{@>}
}

\lstdefinelanguage
   [x64]{Assembler}     
   [x86masm]{Assembler} 
   {morekeywords={CDQE,CQO,CMPSQ,CMPXCHG16B,JRCXZ,LODSQ,MOVSXD, %
                  POPFQ,PUSHFQ,SCASQ,STOSQ,IRETQ,RDTSCP,SWAPGS, %
                  rax,rdx,rcx,rbx,rsi,rdi,rsp,rbp, %
                  r8,r8d,r8w,r8b,r9,r9d,r9w,r9b}} 

\usepackage{listings}
\usepackage{fancyvrb}
\usepackage{color}
\definecolor{lightgray}{rgb}{.9,.9,.9}
\definecolor{darkgray}{rgb}{.4,.4,.4}
\definecolor{purple}{rgb}{0.65, 0.12, 0.82}
\definecolor{commentgreen}{RGB}{63,127,95}

\colorlet{myPurple}{blue!40!red}
\definecolor{myOrange}{RGB}{255,192,0}

\lstdefinelanguage{Solidity}{
  keywords={len,delete,int,void,payable, public, event, contract, typeof, new, true, false, catch, function, return, null, catch, switch, var, if, in, while, do, else, case, break,struct,const,socklen_t,sa_familty_t,char,sockaddr},
  keywordstyle=\color{violet}\bfseries,
  ndkeywords={class, export, boolean, throw, implements, import, this},
  ndkeywordstyle=\color{darkgray}\bfseries,
  identifierstyle=\color{black},
  sensitive=false,
  comment=[l]{//},
  escapeinside={(*@}{@*)},          
  morecomment=[s]{/*}{*/},
  commentstyle=\color{commentgreen}\ttfamily,
  stringstyle=\color{red}\ttfamily,
  morestring=[b]',
  morestring=[b]"
}

\newcommand{\rnum}[1]{\uppercase\expandafter{\romannumeral #1\relax}}

\usepackage{xcolor}

\algnewcommand{\LeftComment}[1]{\Statex \(\triangleright\) #1}

\definecolor{pptbrown}{RGB}{132,60,12}
\definecolor{pptgreen}{RGB}{56,87,35}

\definecolor{pptgrey}{RGB}{202,202,202}
\definecolor{pptgreen1}{RGB}{192,217,170}

\let\OLDthebibliography\thebibliography
\renewcommand\thebibliography[1]{
  \OLDthebibliography{#1}
  \setlength{\parskip}{0pt}
  \setlength{\itemsep}{0pt plus 0.1ex}
}

  \DeclareFontFamily{U}{dutchcal}{\skewchar \font =45}
  \DeclareFontShape{U}{dutchcal}{m}{n}{
    <-> dutchcal-r}{}
  \DeclareFontShape{U}{dutchcal}{b}{n}{
    <-> dutchcal-b}{}
  \DeclareMathAlphabet{\mdutchcal}{U}{dutchcal}{m}{n}
  \SetMathAlphabet{\mdutchcal}{bold}{U}{dutchcal}{b}{n}
  \DeclareMathAlphabet{\mdutchbcal} {U}{dutchcal}{b}{n}

  \DeclareFontFamily{U}{txcal}{\skewchar \font =45}
  \DeclareFontShape{U}{txcal}{m}{n}{
    <-> txr-cal}{}
  \DeclareFontShape{U}{txcal}{b}{n}{
    <-> txb-cal}{}
  \DeclareMathAlphabet{\mtxcal}{U}{txcal}{m}{n}
  \SetMathAlphabet{\mtxcal}{bold}{U}{txcal}{b}{n}
  \DeclareMathAlphabet{\mtxbcal} {U}{txcal}{b}{n}

\newlength{\textfloatsepsave}
\newcommand\freefootnote[1]{%
  \let\thefootnote\relax%
  \footnotetext{#1}%
  \let\thefootnote\svthefootnote%
}

\newcommand{\parh}[1]{\smallskip\noindent\textbf{#1}}

\newcommand{\finding}[2]{
  \smallskip
  \smallskip
\begin{tcolorbox}[width=\linewidth,boxrule=0pt,top=1pt, bottom=1pt, left=1pt,right=1pt, colback=gray!20,colframe=gray!20]
\textbf{Finding #1:} 
{#2}
\end{tcolorbox}}

\usepackage{bbding}
\newcommand{\CBrush}{\textcolor[RGB]{84,130,53}{\Checkmark}}
\newcommand{\XBrush}{\textcolor[RGB]{176,35,24}{\XSolidBrush}}

\begin{document}

\title{An Empirical Study on Large Language Models in Accuracy and Robustness under Chinese Industrial Scenarios}

\author{Zongjie Li}
\affiliation{%
  \institution{Hong Kong University of Science and Technology}
  \city{Hong Kong}
  \country{China}}
\email{zligo@cse.ust.hk}

\author{Wenying Qiu}
\affiliation{%
  \institution{China Academy of Industrial Internet}
  \city{Beijing}
  \country{China}}
\email{qiuwenying@china-aii.com}

\author{Pingchuan Ma}
\affiliation{%
  \institution{Hong Kong University of Science and Technology}
  \city{Hong Kong}
  \country{China}}
\email{pmaab@cse.ust.hk}

\author{Yichen Li}
\affiliation{%
  \institution{Hong Kong University of Science and Technology}
  \city{Hong Kong}
  \country{China}}
\email{ylipf@cse.ust.hk}

\author{You Li}
\affiliation{%
  \institution{China Academy of Industrial Internet}
  \city{Beijing}
  \country{China}}
\email{liyou@china-aii.com}

\author{Sijia He}
\affiliation{%
  \institution{China Academy of Industrial Internet}
  \city{Beijing}
  \country{China}}
\email{hesijia@china-aii.com}

\author{Baozheng Jiang}
\affiliation{%
  \institution{China Academy of Industrial Internet}
  \city{Beijing}
  \country{China}}
\email{jiangbaozheng@china-aii.com}

\author{Shuai Wang}
\authornote{Corresponding author.}
\affiliation{%
    \institution{Hong Kong University of Science and Technology}
    \city{Hong Kong}
    \country{China}}
\email{shuaiw@cse.ust.hk}

\author{Weixi Gu}
\affiliation{%
  \institution{China Academy of Industrial Internet}
  \city{Beijing}
  \country{China}}
\email{guweixi@china-aii.com}
\authornotemark[1]

\begin{abstract}

Recent years have witnessed the rapid development of large language models (LLMs) in various domains. 
To better serve the large number of Chinese users, many commercial vendors in China have adopted localization strategies, training and providing local LLMs specifically customized for Chinese users. 
Furthermore, looking ahead, one of the key future applications of LLMs will be practical deployment in industrial production by enterprises and users in those sectors.
However, the accuracy and robustness of LLMs in industrial scenarios have not been well studied. In this paper, we present a comprehensive empirical study on the accuracy and robustness of LLMs in the context of the Chinese industrial production area. We manually collected 1,200 domain-specific problems from 8 different industrial sectors to evaluate LLM accuracy. Furthermore, we designed a metamorphic testing framework containing four industrial-specific stability categories with eight abilities, totaling 13,631 questions with variants to evaluate LLM robustness.
In total, we evaluated 9 different LLMs developed by Chinese vendors, as well as four different LLMs developed by global vendors. Our major findings include: (1) Current LLMs exhibit low accuracy in Chinese industrial contexts, with all LLMs scoring less than 0.6. Global LLMs excel in reasoning and open-ended tasks, outperforming local LLMs that are better at understanding Chinese terminology. (2) The robustness scores vary across industrial sectors, and local LLMs overall perform worse than global ones. (3) LLM robustness differs significantly across abilities. Global LLMs are more robust under logical-related variants, while advanced local LLMs perform better on problems related to understanding Chinese industrial terminology.
Our study results provide valuable guidance for understanding and promoting the industrial domain capabilities of LLMs from both development and industrial enterprise perspectives. The results further motivate possible research directions and tooling support.

\end{abstract}

\maketitle

\section{Introduction}
\label{sec:intro}

Recent studies have demonstrated that large language models (LLMs) exhibit exceptional performance across various natural language processing tasks, rivaling or even exceeding human competencies in certain areas~\cite{brown2020language,radford2019language,wang2023reef,ma2023insightpilot}. Typically, LLMs undergo pre-training on extensive text corpora, usually using billions of tokens to develop a foundational model~\cite{chen2021evaluating,brown2020language,touvron2023llama2,wei2023skywork}. To better align LLMs with human preferences and directives or to fulfill specific application needs, methods such as supervised fine-tuning (SFT), reinforcement learning from human feedback (RLHF)~\cite{ouyang2022training}, and direct preference optimization (DPO)~\cite{rafailov2023direct} have been introduced and demonstrated to be effective. These advancements facilitate more intuitive and efficient human-AI interactions. 
However, the substantial resource requirements throughout the training process pose challenges for individual users and smaller organizations.

Despite the success of LLMs, most are trained with English corpora, and the majority of LLMs are designed for English users.
For example, prominent open-source LLMs such as LLaMA2~\cite{touvron2023llama2} are predominantly trained on English corpora with limited ability in understanding non-English texts, while advanced multilingual models like GPT-4~\cite{gpt4} are often proprietary and unavailable publicly. 
Therefore, to cater to the Chinese market and adhere to legal regulations, many Chinese LLM vendors adopt localization strategies, providing ``local'' LLM specially designed for Chinese by pretraining models on large-scale Chinese datasets~\cite{chatglm2, CLUECorpus2020, yuan2021wudaocorpora} or fine-tuning them using supervised instruction corpora~\cite{redgpt,li-etal-2022-csl}.

In recent times, there has been a growing trend of integrating LLMs into manufacturing production pipelines~\cite{palmsec,maatouk2023large}. However, the high threshold for training resources and corpora, along with the lack of testing for non-English LLMs~\cite{zhu2023extrapolating}, raises concerns regarding their accuracy and robustness. 
In the context of industrial manufacturing, accuracy is crucial to prevent potential catastrophic defects that could result in significant losses~\cite{bard-lose}. 
Additionally, robustness is vital as manufacturing models frequently operate under constrained conditions, necessitating deterministic outputs rather than conversational ones. Despite the successes in dialogue applications, industrial manufacturing organizations remain hesitant to adopt LLMs within critical production systems due to these challenges. Ensuring sufficient accuracy and reliability to gain stakeholder trust and mitigate risks remains a significant hurdle for the widespread implementation of LLMs in the manufacturing sector.

Currently, we notice that several benchmarks exist for evaluating local LLMs~\footnote{In this study, we use local LLM to refer to Chinese local LLM unless otherwise specified.}, like dialogue ability~\cite{FlagEval}, commonsense reasoning~\cite{huang2023ceval}, and creative writing~\cite{2023opencompass, Zhang2023EvaluatingTP, zhong2023agieval,li2023cmmlu,zeng2023measuring}. However, these focus on conversational settings, assessing skills like poetry generation that lack industrial relevance. Hence, specialized studies evaluating local LLM accuracy and robustness for manufacturing applications remain scarce. As a result, there is a need for specialized studies that evaluate local LLMs' accuracy and robustness in manufacturing applications.

To address this gap, we present a comprehensive empirical study that assesses the accuracy and robustness of local LLMs in industrial scenarios. The study was conducted through collaboration between nine top industrial research teams in China. We carefully considered relevant administrative regulations and laws and manually curated 1,200 industry-specific problems across eight industrial sectors to assess accuracy.
To further quantify the robustness of the LLMs, we developed a metamorphic testing framework with industry-oriented relations. This framework evaluated four stability categories with eight abilities through 12,431 variants. In total, we evaluated eight different local LLMs developed by Chinese vendors and two different LLMs developed by global vendors.
Overall, our study aims to address the following four research questions (RQs):

\textbf{RQ1:} To what extent are LLMs accurate in Chinese industrial scenarios?

\textbf{RQ2:} How robust are LLMs across different Chinese industrial scenarios?

\textbf{RQ3:} What differences exist in LLM robustness capabilities across industrial sectors?

\textbf{RQ4:} How does local LLM's robustness vary across diverse robustness abilities?

We obtain many findings, and some of the major ones include:

\begin{itemize}
  \item The accuracy of all evaluated LLMs remains insufficient (under 60\%) for deployment in industrial applications, and it varies substantially across industrial areas, especially for less mature models.
  \item Most LLMs tend to be more robust on the variants whose corresponding seed question is correctly answered than those answered wrongly.
  \item The robustness scores vary across industrial sectors, and local LLMs are generally worse than global LLMs, which may be attributed to the quantity and quality of available data on the Internet.
  \item LLM robustness differs significantly across abilities. Global LLMs demonstrate greater robustness to logical perturbations, while top local LLMs better understand Chinese industrial terminology.
  
\end{itemize}

Our findings provide guidance for developing LLMs that better serve non-English (Chinese) users in industrial applications. They will assist platform engineers and enterprises in improving local LLMs for manufacturing. 

To summarize, this paper makes the following contributions:
\begin{itemize}
  \item We perform the first comprehensive study on the accuracy and robustness of LLMs in Chinese industrial scenarios.
  \item We collect the first benchmark of industry-specific problems in Chinese.
  \item We propose a metamorphic testing framework with industrial metamorphic relations to assess robustness in Chinese industrial scenarios.
  \item We systematically evaluate 10 LLMs from 9 different vendors, and compare the local LLMs with global LLMs in terms of multiple abilities.
  \item We point out the implications of our findings and suggest possible improvements for the development and usage of LLMs in Chinese industrial scenarios.
\end{itemize}

The rest of the paper is organized as follows.
In~\S~\ref{sec:data-collection}, we introduce industrial scenarios considered and criteria for collecting industry-specific benchmarks.
\S~\ref{sec:design} presents the metamorphic testing framework with industry-specific metamorphic relations for different abilities.
\S~\ref{sec:implementation} describes the experimental setup and details.
In~\S~\ref{subsec:rq1} and~\S~\ref{subsec:rq2}, we describe the overall results in terms of accuracy and robustness, respectively.
\S~\ref{subsec:rq3} and~\S~\ref{subsec:rq4} introduce our findings on how the different industrial sectors and abilities affect the robustness of LLMs.
We survey related work in~\S~\ref{sec:related} and discuss the threats to validity and possibility of extension in~\S~\ref{sec:discussion}.
Finally, we conclude in~\S~\ref{sec:conclusion}.

\section{Industrial Data Collection}
\label{sec:data-collection}

\subsection{Key Industrial Sectors}
\label{subsec:key-sectors}

\begin{table*}[!htpb]
	\caption{Four categories and eight abilities of the stability category. }
    \vspace{-10pt}
	\begin{tabular}{lllll}
	\toprule
	Name & Stability  & Description & Original texts & Equivalent variants \\  \midrule
	Magnitude change     & Numeric     &    Equivalent substitution of data outlines         &  100cm    &   1m    \\
	Digital precision	 & Numeric        &  Changing data precision           &  3.7m     &  3.70m     \\
	Synonyms  & Grammar     	  &    Replacement with industry-specific synonyms    &  USB      &   U-PAN    \\
	Order     & Grammar     &    Swapping the order of options          &  A. five  B. six     &   A. six  B. five    \\
	Logic     & Grammar     &     Reversal the logic of the question        &  You should touch xx     & You should not touch xx      \\
	General context	 & Context    &  Add background for testing Industrial sectors           &  [Q]  &  In xxx industry, [Q]     \\
	Security context & Context               &  Add security instruction          &  [Q]     &   Consider xxx law, [Q]    \\
	Irrelevant content & Others    &   Add irrelevant option   &   A|B|C|D    &  A|B|C|D|E     \\  \bottomrule
	\end{tabular}
    \vspace{-10pt}
	\label{tab:stability-category}
\end{table*}

The industrial base of a mature country typically comprises thousands of sectors across various industries such as manufacturing, mining, energy, and materials~\cite{bernstein2007industry}. China, for instance, boasts one of the most diverse and extensive industrial bases globally, with tens of thousands of sectors~\cite{chineseindus}. Among these numerous sectors, eight key industries can be identified that have high production values, utilize advanced automation, and have the potential for further integration of LLMs into their industrial chains. These industries are:

\begin{itemize}

  \item Electronic equipment manufacturing: Fabrication of electronic devices, components, specialized materials, and other electronic components.
  
  \item Equipment manufacturing: Production of metal products, general equipment, specialized equipment, and automobiles.
  
  \item Iron and steel industry: Ironmaking, steelmaking, steel rolling and processing, ferroalloy smelting.
  
  \item Mining industry: Extraction of coal, oil, natural gas, ferrous metals, non-ferrous metals, and other minerals.
  
  \item Power industry: Electricity generation, transmission, heat production, and distribution.
  
  \item Petrochemical industry: Petroleum refining and processing, manufacturing of chemical feedstocks and products, plastic goods, rubber goods.
  
  \item Building materials industry: Manufacturing of construction materials and products, non-metallic minerals and products, inorganic non-metallic novel materials.
  
  \item Textile industry: Cotton, wool, linen, silk, synthetic fibers and other textiles, printing and dyeing finishing processes.
  
\end{itemize}

The industrial sectors examined represent fundamental components in the manufacturing chain, as well as the industries with the most extensive real-world applications. These sectors are selected due to their importance in industrial processes and ubiquity across supply chains. \textbf{Notably, the research focuses exclusively on civilian industries and does not encompass any sectors related to national security or defense.}

\subsection{Data Origin}
\label{subsec:dataset}

As outlined in~\S~\ref{sec:intro}, the existing benchmarks for LLM evaluation are primarily relied on open-source datasets, which often consist of conversational logs or focus on common sense knowledge. 
However, given that the industrial chain operates as a closed-loop system, it may be challenging to persuade stakeholders that the data industrial sectors need is from open-ended conversations rather than well-structured, formal, and professional documents.
Therefore, we collect the data from three real-world sources, as follows:

\textbf{National Authoritative Question Set.}~It includes test questions used in vocational re-education tests to check whether the examinee has a strict grasp of the production norms of the relevant industry.

\textbf{National vocational and technical training guidelines.}~This material is intended as a general guide for undergraduate and graduate students who intend to pursue a career in a related field.

\textbf{Production and operation standardization conditions.}~For different industries, corresponding functional government agencies have issued relevant regulatory guidelines to ensure that industrial production is carried out in safe and controlled conditions, and this part of the material is more targeted and detailed than the previous two.

We notice that different countries have varying requirements and regulations that are dependent on their level of industrialization and local laws. Therefore, we claim that we only collect the data from the industrial sectors in China as this is the research scope of this paper. However, we believe that our method can be applied to other countries, provided  the data that is collected from qualified industrial documents conforming to the local laws and regulations.

\subsection{Question Design}
\label{subsec:question-format}

In this section, we first introduce the question formats used in this study, and then give the corresponding accuracy definitions.

\parh{Question Format.}~After collecting a sufficient set of problems, we organize the data into formats appropriate for evaluating LLMs. Following prior work like CMMLU~\cite{li2023cmmlu}, we primarily adopt multiple-choice questions. 
Furthermore, we incorporate true/false questions since many of the collected problems from our data source are in this format. These formats have ground truth answers, making evaluation straightforward. As we mentioned in~\S~\ref{sec:intro}, existing LLM benchmarks also use open-domain conversational logs. Although these open-ended questions are less prioritized in industrial applications, we still include them in our evaluation to provide a comprehensive assessment of LLMs.

\parh{Accuracy Definition.}~Following prior work~\cite{li2023split}, we use varying definitions of accuracy for different question formats. For the questions that have ground truth answers, we calculate the accuracy by dividing the answers that are correct by the total number of all tested questions to evaluate the performance of the LLMs. 
On the other hand, for the open-ended question, we recruit twenty experts as participants. All the participants are fully experienced experts in specific industrial sectors.  
Each participant is provided with an online questionnaire containing one question with one response, without specifying the origin. Before the questionnaire, we conducted a tutorial course to ensure fair and objective assessment. The assessment criteria are mainly based on National Higher Education Entrance Examination in China. For any disagreements, we applied majority voting to determine the final result. In summary, the final LLM accuracy is determined as the weighted average of the multiple-choice, yes/no, and open-ended questions.

\section{Metamorphic Framework design}
\label{sec:design}

As discussed in~\S~\ref{sec:data-collection}, we first collect 1,200 seed questions from three different sources. 
Then, we use metamorphic testing (MT)~\cite{chen1998metamorphic} to systematically generate the MT variants of the questions. Specifically, We categorize the different transformation methods into four classes according to the stability capability they are intended to test, and transform them accordingly.
Finally, we use the generated variants to evaluate the robustness of LLMs.

\parh{Stability Category.}~The stability category focuses on an in-depth evaluation of LLMs along eight abilities across four categories: numeric, grammatical, contextual, and others. \T~\ref{tab:stability-category} outlines the eight abilities, providing their names and descriptions, as well as examples of original texts and their equivalent variants.
The numeric category examines LLMs' capability to handle numeric data, including changes in magnitude and precision.
On the other hand, the grammatical category, comprising the ``Synonyms'', ``Order'', and ``Logic'' abilities, evaluates LLMs' ability to handle grammatical and semantic changes in questions.
In contrast to the previous two categories focused on token-level replacements, the context and other categories concentrate on sentence-level transformations.
As the name suggests, the context category assesses LLMs' stability under different contexts, including general and security-related contexts. 
Additionally, the sole ability in the others category handles the inclusion of irrelevant content in questions.

\parh{Robustness Definition.}~The robustness examined here refers to an LLM's ability to maintain performance when facing input prompt variations in industrial applications. Unlike accuracy evaluations which include open-ended questions as inputs, our robustness evaluation only considers multiple-choice and true/false questions as the original inputs, since open-ended questions are exceedingly difficult to evaluate automatically without bias~\cite{zheng2023judging}. 
Robustness is defined as the ratio of variants that share the same answer as the seed question, with a higher score indicating better robustness.

\section{Implementation and Setup}
\label{sec:implementation}

\begin{table}[!htbp]
    \centering
    \caption{All LLMs in our consideration. ``Deploy'' denotes the deployment mode of the LLMs. ``Chosen ?'' denotes whether the LLMs are chosen for evaluation.}
    \resizebox{1.00\columnwidth}{!}{
    \begin{tabular}{l|c|c|c|c}
    \hline
    \textbf{Model} & \textbf{Vendor}   & \textbf{Version} & \textbf{Deploy} & \textbf{Chosen ?} \\ \hline
    GPT-4~\cite{gpt4}            & OpenAI      & gpt-4-0613& API   &  \CBrush\\ \hline
    GPT-3.5~\cite{ChatGPT}       & OpenAI      & gpt-3.5-turbo-0301& API   &  \CBrush\\ \hline
    Tongyi~\cite{qwen}      	 & Alibaba     &  v1.0.5& API   &  \CBrush\\ \hline
    Ernie~\cite{ernie}           	 & Baidu       &v2.2.2& API   &  \CBrush\\ \hline
    spark~\cite{sparkapi}   	 & Iflytek     & v2.0	& API   &  \CBrush\\ \hline
    ChatGLM2~\cite{chatglm2}	 & ZHIPU       &  6b  & Local &  \CBrush\\ \hline
    Minimax~\cite{minimax}       & Minimax     &  v5.5 & API   &  \CBrush\\ \hline
    Tiangong~\cite{Tiangong}     & Kunlun      &v3.5.20230705.a	& WEB   &  \CBrush\\ \hline
    Congrong~\cite{CongRong}     & Cloudwalk   & v20230518	& WEB   &  \CBrush\\ \hline
    360ZhiNao~\cite{360zhinao}   & 360         & v9.5 		  & API   &  \CBrush\\ \hline \hline
    Llama2~\cite{touvron2023llama2}& Meta AI   & 13b   & Local &  \XBrush\\ \hline
  	Claude2~\cite{claude2}       & Anthropic   & claude2   & API   &  \XBrush\\ \hline
    \end{tabular}    
    }
	
    \label{tab:testedllm}
\end{table}

\parh{Tested LLMs.}~As shown in~\T~\ref{tab:testedllm}, we evaluate a total of 12 LLMs, including 4 global LLMs and 8 local LLMs. 
The ``Tiangong'' and ``Congrong'' models are used through their respective web services to generate question answers, while other models are accessed through API services or locally deployed if source code is available. However, we exclude results from Llama2 and Claude2 in our primary evaluation, as they perform unsatisfactorily with an accuracy rate of less than 0.1.
Unfortunately, most vendors do not provide information on the number of parameters and the training dataset size of each LLM, hence we could not provide such details in our evaluation. Among all the LLMs, we only know that GPT-3.5 used to have 175 billion parameters, ChatGLM2 is a 6 billion parameters model, and llam2-13b is a 13 billion parameters model. 
Notably, comparing the GPT series models with local LLMs may not be entirely fair, given the lack of clarity on these details. Nevertheless, since the GPT models are considered state-of-the-art in the general domain and the local LLMs are primarily designed for Chinese users, 
we believe that the comparison still provides valuable insights into the current state of LLMs in the industrial domain.

\parh{Dataset Statistics.}~The original dataset used in this study contains 1,200 questions evenly divided across 8 industrial sectors, with each sector contributing 150 questions. 
As outlined in~\S~\ref{sec:design}, our MT framework generates variants across four stability categories with eight abilities.
However, the number of generated variants varied for each ability due to certain metamorphic relations being incompatible with specific question formats. For instance, the ``Order'' relation can not be applied to yes/no questions as they lack options. Similarly, relations involving numerical value modification can only be applied to questions containing numbers, which are not evenly distributed across sectors. This imbalance is retained to reflect real-world conditions and the nature of the different industrial sectors.

Finally, the variant dataset we use for robustness testing contains a total of 13,631 questions, including 12,431 variants and 1,200 original questions.
\T~\ref{tab:dataset} shows the number of variants generated for each sector, where we see that the least number of variants generated for each sector is more than one thousand, which is sufficiently large to facilitate a thorough assessment of the LLMs' robustness.

\begin{table}[!htbp]
	\centering
  \caption{The statistics of variants generated for each sector. ``\# Variants'' denotes the number of variants generated for each sector.}
	\label{tab:dataset}
		\begin{tabular}{l|l|l|l}
			\hline
			\textbf{Industry}  & \textbf{\# Variants} & \textbf{Industry} &\textbf{\# Variants}\\ \hline
			Electronic         & 1,479   & Power & 1,955 \\
			Equipment          & 1,843   & Petrochemical & 1,060 \\
			Iron               & 1,377   & Building & 1,055 \\
			Mining             & 1,254   & Textile & 1,208 \\ \hline
			\end{tabular}
\end{table}

\parh{Experiments Setup.}~During the experiments, we observe that LLMs can exhibit instability, as multiple attempts with the same prompt may produce diverse results.
This is primarily due to the inherent randomness of the LLMs' decoding process, in which tokens are sampled from a probability distribution over the vocabulary. While this stochasticity enhances response diversity in open-domain conversation systems, consistency is crucial for industrial applications.
Consequently, we enforce determinism to guarantee replicable evaluations and facilitate reliable comparisons. To minimize randomness and ensure reproducibility in our experiments, we employ various methods. For cloud API-based models, we assigned a value of 0 to the ``temperature'' hyper-parameter. In the case of locally deployed models, we deactivate sampling during decoding to achieve deterministic outcomes. However, controlling the hyper-parameters of LLMs via a web service is unfeasible due to the absence of a suitable interface. As a solution, we perform multiple experimental runs and select representative results using a majority vote approach.
Additionally, we also notice that for GPT models, the results would be different even if we turn the ``temperature'' to 0, which is in line with the observation in~\cite{ouyang2023llm}.
All experiments are conducted on a machine with the Intel Xeon Platinum 8276 CPU, 256 GB main memory and 4 NVIDIA A100 GPUs.
\section{Evaluation}
\label{sec:evaluation}

In accordance with the four research questions (RQs) noted in \S~\ref{sec:intro}, we demonstrate the results and answer each of them in this section. 
Due to the confidentiality agreements with vendors, we anonymize the names of all LLMs except the top six highest accuracy models. The anonymized models are referred to as AN\_Model 1, AN\_Model 2, and so on.

\subsection{RQ1: LLM Accuracy}
\label{subsec:rq1}

\begin{figure}[!htpb]
    \centering
    \includegraphics[width=0.45\textwidth]{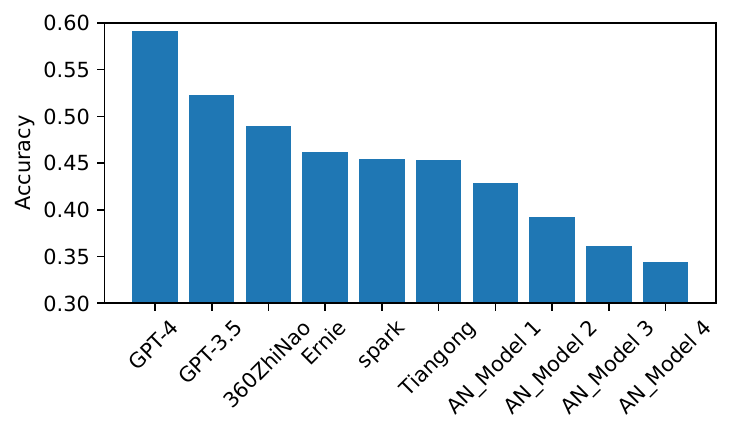} 
    \caption{The accuracy of LLMs in the Chinese industrial scenarios.}
    \vspace{-10pt}
    \label{fig:acc}
\end{figure}

This section examines the overall accuracy of LLMs in Chinese industrial scenarios. \F~\ref{fig:acc} presents the results, where the value of accuracy is the average value of a given LLM across all eight industrial sectors. 
From the figure, we can see that the accuracy rates of LLMs are concerningly low at less than 0.6. 
This is far inferior to reported accuracy on general domain benchmarks. For instance,~\citet{brown2020language} report GPT-3.5 achieving over 86\% on the LAMBADA language modeling dataset, while~\citet{huang2023ceval} find GPT-4 attains over 74\% on social science reasoning benchmark of C-eval. 
Moreover, we observe a significant disparity in the accuracy rates of various LLMs. Notably, the highest accuracy rate of 0.59 is attributed to GPT-4, while ``AN\_Model 4'' records the lowest accuracy rate of 0.33, indicating a substantial gap among the LLMs. Furthermore, the standard deviation of 0.2, which is considerably high, underscores the considerable variability of the accuracy.

When examining only the top six highest accuracy LLMs, we surprisingly find that the accuracy of local LLMs is overall lower than global ones in Chinese industrial scenarios. To better understand this gap, we manually reviewed 10 answers from each industrial sector and made two key observations: (1) Generally speaking, GPT series models demonstrate stronger capabilities in understanding questions related to mathematics, while local LLMs are more adept at comprehending questions involving special terms in Chinese. (2) For open-ended questions, GPT series models tend to provide answers with clearer logic, which results in higher accuracy ratings from human evaluators.
These findings indicate that despite having access to more Chinese training data, current local LLMs still lag behind global models, especially GPT-4, in many industrial applications. While local LLMs may have some advantages in parsing Chinese vocabulary, their reasoning and open-ended generation abilities require improvement to reach parity with global leaders like GPT-4.

\begin{figure}[!htpb]
    \centering
    \includegraphics[width=0.45\textwidth]{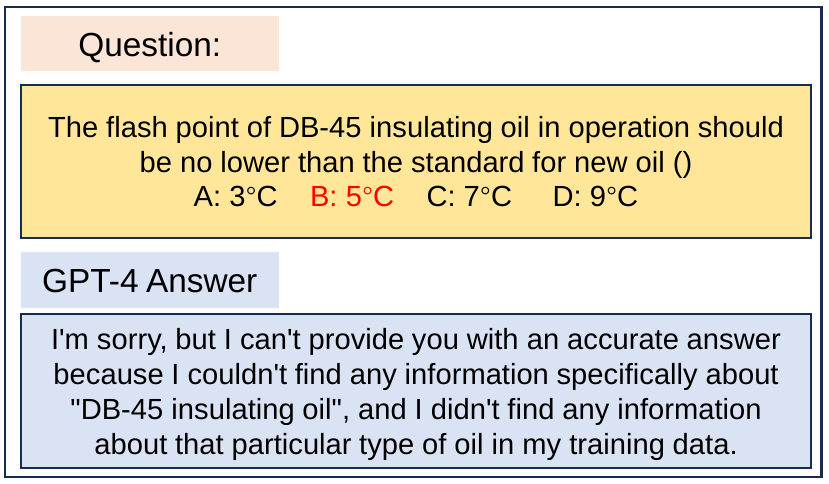} 
    \caption{The example of the self-knowledge blind spot discovery of GPT-4. The correct answer is B, marked in red.}
    \label{fig:self-blind}
\end{figure}

\parh{Identifying Knowledge Blind Spot.}~An intriguing finding is the potential for LLMs like GPT-4 to actively demonstrate when a question falls outside their knowledge scope. Instead of providing an incorrect answer, acknowledging the knowledge limitations could prevent misleading users.
For example, in~\F~\ref{fig:self-blind}, GPT-4 responds that it lacks the requisite information on ``DB-45 insulating oil'' in its training corpus to answer the question. While it is uncertain if GPT-4 truly lacks this knowledge or is just deceiving the user, admitting a knowledge gap is arguably preferable to supplying a faulty response. This self-awareness of blind spots can rapidly help organizations identify areas for improvement in industrial applications of LLMs. Moreover, this behavior is not exclusive to GPT models; some local LLMs like ChatGLM2 and Ernie occasionally demonstrate similar knowledge limitation admissions, albeit less frequently. Overall, we view the ability to acknowledge the boundaries of installed knowledge represents a promising development for responsible and transparent LLM deployment.

\finding{1}{Current LLMs exhibit low accuracy in Chinese industrial contexts, with all models scoring less than 0.6. Further human evaluation reveals that global LLMs excel in reasoning and open-ended tasks, outperforming local LLMs that are better at understanding Chinese terminology. Additionally, some LLMs can identify their knowledge blind spot, which is promising for responsible and transparent LLM deployment.}

\subsection{RQ2: Overall Robustness}
\label{subsec:rq2}

In this section, we evaluate the robustness of LLMs based on two categories: ``correct robustness'' and ``incorrect robustness''.
Given an original seed question, if a particular LLM provides an erroneous reply, the robustness score calculated by this seed as well as its variants is categorized as ``incorrect robustness.''
Conversely, if it produces a correct reply, we classify it as ``correct robustness.''
By examining both correct and incorrect LLM responses, we aim to provide a comprehensive perspective on overall LLM robustness.
However, it is important to note that our analysis beyond this section focuses solely on ``correct robustness.'' This is because for industrial applications, answer correctness is the main priority, and stable output generation is more important than analyzing incorrect responses. 
Therefore, we only examine the ``correct robustness'' in~\S~\ref{subsec:rq3} and~\S~\ref{subsec:rq4}.

\begin{table}[h]
    \centering
    \begin{tabular}{|l|c|c|}
    \hline
    Model & Correct Robustness & Incorrect Robustness \\
    \hline
    \textbf{GPT-4} & 0.76016 & 0.75595 \\
    \textbf{GPT-3.5} & 0.71681 & 0.69633 \\
    \textbf{Ernie} & 0.62683 & 0.57975 \\
    \textbf{spark} & 0.70407 & 0.72076 \\
    \textbf{Tiangong} & 0.70007 & 0.43804 \\
    \textbf{360ZhiNao} & 0.70694 & 0.55201 \\
    \textbf{AN_Model 1} & 0.63711 & 0.58077 \\
    \textbf{AN_Model 2} & 0.49817 & 0.38892 \\
    \textbf{AN_Model 3} & 0.65318 & 0.62105 \\
    \textbf{AN_Model 4} & 0.63128 & 0.63054 \\
    \hline
    \end{tabular}
    \caption{Robustness scores for various models.}
    \vspace{-10pt}
    \label{tab:robustness}
    \vspace{-10pt}
\end{table}

The results presented in~\T~\ref{tab:robustness} demonstrate that the robustness scores for correct and incorrect answers generally follow similar trends across most LLMs, with the correct robustness consistently exceeding the incorrect robustness. One exception is the ``spark'' model, which displays higher incorrect robustness than correct robustness. 
It is important to note that during the test, we initially used version 1.0 of the ``spark'' model, which had unsatisfactory robustness scores for both correct and incorrect answers. We reported this issue to the vendor, and they updated the model to version 2.0, which showed a significant improvement in robustness, especially for incorrect answers. As a result, we believe that the ``spark'' model may be trained on some adversarial examples or given some pre-defined rules, which enhanced its robustness for incorrect answers.
Furthermore, we find that correct robustness scores strongly correlate with LLM accuracy. For instance, GPT-4 has the highest correct robustness score at 0.76 and also the highest accuracy among the LLMs. In contrast, ``AN_Model 2'' has the lowest correct robustness score at 0.49 and the lowest accuracy. 

It is worth noting that during the experiments, we find that some of the tested LLMs would decline to answer certain questions, particularly those related to enterprise security.
For example, when we ask some local LLMs ``According to the National Standard of the People's Republic of China, the General Principles of Design for Safety and Protection of Production Equipment, the integrated operational amplifier is essentially a multistage direct-coupled amplifier.'', they would refuse to provide an answer.
However, if we simply delete the half part of the question, the model would give the meaningful answer ``True''.
We further check the status code in the response of the model, and find that some of the LLMs such as GPT-3.5 and MiniMax would return a special status code to demonstrate that the question is rejected due to security concerns. 
In such cases, we exclude the question from the robustness evaluation, and the LLM's robustness score would not be affected by this variant.

\finding{2}{The correct robustness scores of LLMs are generally higher than their incorrect robustness scores, indicating that industrial LLM usage prioritizes answer correctness over robustness. Additionally, correct robustness strongly correlates with LLM accuracy.}

\subsection{RQ3: Robustness Across Sectors}
\label{subsec:rq3}

\begin{figure*}[!htpb]
    \centering
    \includegraphics[width=1.0\textwidth]{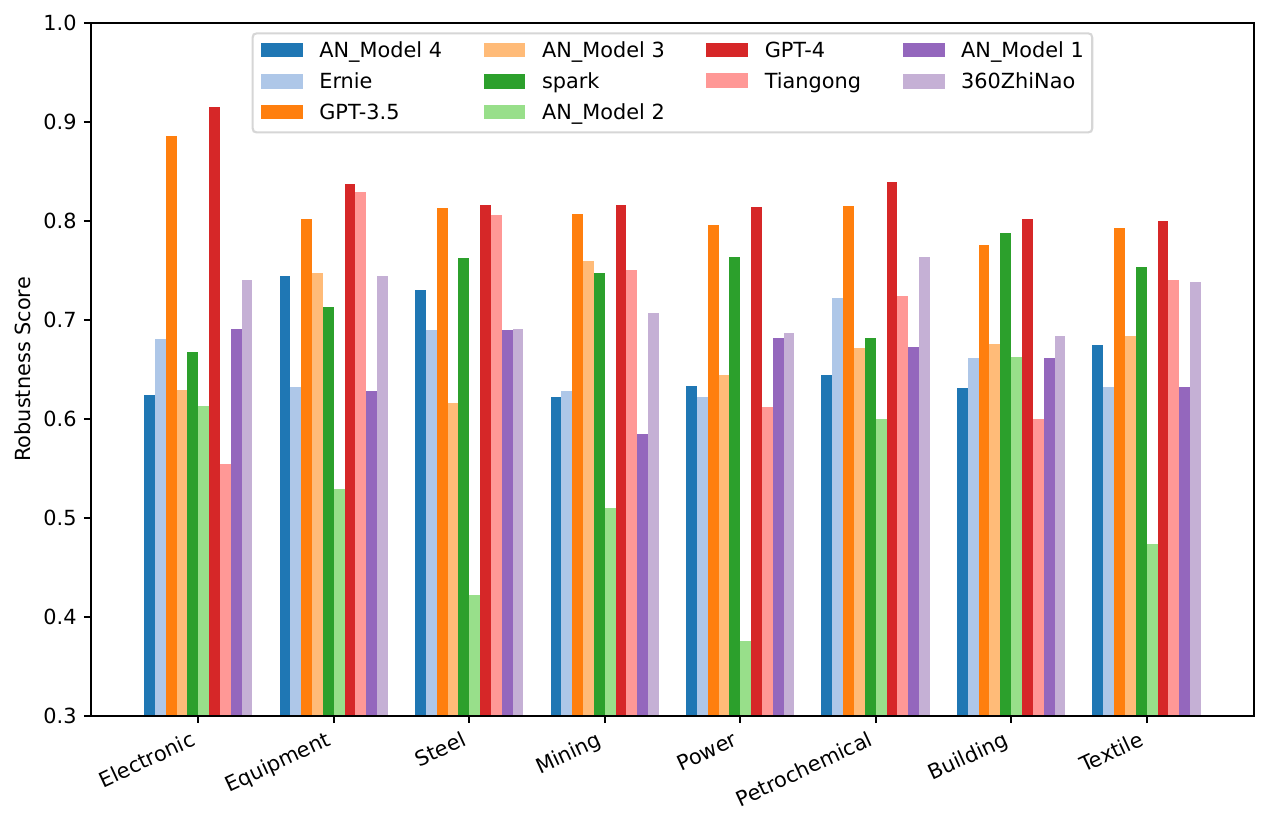} 
    \vspace{-15pt}
    \caption{The robustness scores for different LLMs on various industrial sectors.}
    \vspace{-10pt}
    \label{fig:robust1}
\end{figure*}

This section investigates the robustness of LLMs across different industrial sectors. The results are presented in \F~\ref{fig:robust1}, where the ``Robustness score'' represents the average robustness value of a specific LLM across all proposed abilities. In this section, we only consider "correct robustness," as discussed in ~\S~\ref{subsec:rq2}.

To begin, we focus on the performance of the same LLM in different sectors. 
From the figure, we can observe that for one specific LLM, the robustness scores in different industrial sectors follow a similar trend. 
In other words, an LLM is unlikely to achieve the highest robustness in one sector but the lowest robustness in another sector. 
This indicates that the robustness of LLMs is highly dependent on their architecture and parameters, and well-trained LLMs are more likely to achieve high robustness across all industrial sectors.
Furthermore, we observe that the robustness scores of LLMs vary across different industrial sectors. For instance, the ``Power'' sector has the lowest average robustness score of 0.66, while the ``Equipment'' sector has the highest score of 0.72. Additionally, we find that LLMs provided by local vendors generally have lower robustness scores than those from global vendors. For example, in the ``Electronic'' sector, the average robustness score of local LLMs is 0.65, while global LLMs have a score of 0.90. Even when considering only the top 3 local LLMs with the highest robustness score, the average robustness score gap is still around 0.13.

\begin{table}[h] 
    \centering 
    \caption{Related information for industrial sectors on Baidu and Google. \# Items represents the number of related items returned by the search engine, and Rank represents the rank of the number of related items among all industrial sectors.} 
    \label{tab:search} 
    \begin{tabular}{|l|c|c|c|c|} \hline \multicolumn{1}{|c|}{\multirow{2}{*}{Industries}} & \multicolumn{2}{c|}{Baidu} & \multicolumn{2}{c|}{Google} \\ \cline{2-5} \multicolumn{1}{|c|}{} & \multicolumn{1}{l|}{\# Items} & \multicolumn{1}{l|}{Rank} & \multicolumn{1}{l|}{\# Items} & \multicolumn{1}{l|}{Rank} \\ \hline 
        Electronic &54,200,000 & 7& 56,700,000 & 1 \\ \hline 
        Equipment & 82,000,000 & 5&24,500,000 & 3\\ \hline 
        Steel &90,900,000 & 4 &22,700,000 & 4\\ \hline 
        Mining &55,600,000 &6 &10,300,000 & 7\\ \hline 
        Power &100,000,000 & 1 &33,000,000 & 2 \\ \hline 
        Petrochemical &47,200,000 & 8 &7,660,000 & 8\\ \hline 
        Building &100,000,000 & 1 &18,300,000 & 6\\ \hline 
        Textile &95,500,000 & 3 &19,000,000 & 5\\ \hline
    \end{tabular} 
    \vspace{-10pt}
\end{table}

\parh{Industrial Information Analysis.}~To gain a better understanding of the underlying reasons, we perform a statistical analysis of industry-specific information available on the internet. Our approach involves using both Baidu and Google search engines to gather data on relevant topics in \textbf{Chinese}, and subsequently tallying the number of results obtained. Since these search engines are widely used in China and globally, 
we consider them to be an indicator of the amount of relevant information on the Internet.

The results are shown in~\T~\ref{tab:search}, where we can observe that except for the ``Electronic'' sector, the number of items returned by the Baidu search engine is much higher than the number of items returned by the Google search engine. 
We attribute this to the fact that the Baidu search engine is more popular in China, and the related information on the Chinese web is more likely to be indexed by Baidu.

Moreover, we can conclude with two interesting observations. 
First, we find that there is a positive correlation between the number of search results returned by Google and the robustness of the GPT series. Specifically, the ``Electronic'' sector has the highest number of search results, and GPT-4 and GPT-3.5 have the highest robustness scores in this sector among all. We believe that this is because the GPT series are trained on a large corpus from the internet, and the more relevant information available on the internet, the better the GPT models can understand industrial scenarios, which in turn enhances their robustness.

Second, we observe that the robustness of LLMs in Chinese industrial scenarios has a negative correlation with the number of items returned by the Baidu search engine. Surprisingly, the ``Power'' and ``Building'' sectors, with the highest number of search results, have relatively low robustness scores among all sectors. 
This does not comply with our first intuition that more information improves the robustness score. Therefore, we further investigate the returned items from the Baidu search engine, and find that most of the returned items are not directly related to the industrial scenarios, but are related to some general knowledge or outreach. 
Therefore, we attribute this phenomenon to the mismatch between the questions in our benchmark and retrieved information.
Specifically, the questions in our benchmark are all from real industrial scenarios, and correctly answering the questions not only requires the LLMs to understand the general knowledge about the specific industrial sector, but also asks for high-quality corpus to be collected and trained with the LLMs.

Based on the analysis of the related items on the Internet, it can be inferred that the quantity and quality of the retrieved data used for training have a significant impact on the robustness of LLMs. This may result in a disparity between local and global LLMs, as well as between various industrial sectors.

\finding{3}{The robustness of LLMs in different industrial sectors is highly related to the architecture and parameters of LLMs. The robustness scores vary across industrial sectors, and local LLMs overall perform worse than global ones. The disparity between the performance is likely due to differences in the quantity and quality of collected data.}

\subsection{RQ4: Robustness Across Abilities}
\label{subsec:rq4}

\begin{table*}[!htpb]
    \centering
    \caption{Robustness scores of Various LLMs across abilities.}
    \vspace{-10pt}
    \label{tab:performance}
    \begin{tabular}{|l|c|c|c|c|c|c|c|c|}
    \hline
    \textbf{Model} & \textbf{Synonyms} & \textbf{Order} & \textbf{Logic} & \textbf{Digital} & \textbf{Magnitude} & \textbf{Security} & \textbf{General} & \textbf{Irrelevant} \\ \hline
    \textbf{GPT-4}   & 0.9569 & 0.7867 & 0.8625 & 0.8916 & 0.8667 & 0.8734 & 0.9176 & 0.8587 \\ \hline
    \textbf{GPT-3.5} & 0.9419 & 0.7692 & 0.8372 & 0.8814 & 0.8333 & 0.8467 & 0.9017 & 0.8316 \\ \hline
    \textbf{Ernie} & 0.8409 & 0.5596 & 0.6467 & 0.7356 & 0.6364 & 0.7422 & 0.7975 & 0.6224 \\ \hline
    \textbf{spark}   & 0.8814 & 0.6139 & 0.6846 & 0.7980 & 0.7364 & 0.7963 & 0.8680 & 0.7628 \\ \hline
    \textbf{Tiangong} & 0.7867 & 0.6417 & 0.7000 & 0.6667 & 0.3333 & 0.8000 & 0.7671 & 0.7882 \\ \hline
    \textbf{360ZhiNao} & 0.8728 & 0.6467 & 0.6537 & 0.7699 & 0.6875 & 0.8342 & 0.8247 & 0.7075 \\ \hline
    \textbf{AN_Model 1} & 0.9016 & 0.5327 & 0.4731 & 0.8448 & 0.8000 & 0.7803 & 0.8446 & 0.6940 \\ \hline
    \textbf{AN_Model 2} & 0.6837 & 0.5003 & 0.7063 & 0.4545 & 0.5000 & 0.5844 & 0.6074 & 0.4623 \\ \hline
    \textbf{AN_Model 3} & 0.8889 & 0.5997 & 0.5361 & 0.6600 & 0.4706 & 0.7847 & 0.7981 & 0.7340 \\ \hline
    \textbf{AN_Model 4} & 0.9686 & 0.4555 & 0.4552 & 0.8646 & 0.9167 & 0.8252 & 0.8858 & 0.7574 \\ \hline
    \end{tabular}
    \vspace{-10pt}
\end{table*}

This section primarily examines the robustness of LLMs across various abilities. As introduced in~\S~\ref{sec:design}, we propose eight MT relations to evaluate LLMs' robustness, with each relation designed to assess robustness within a specific ability. For an original seed question and its variants, we only test variants whose corresponding original was answered correctly by the model (to as ``correct robustness'' in~\S~\ref{subsec:rq2}).

The results are presented in~\T~\ref{tab:performance}, where the ``Robustness score'' denotes the average robustness value for a given model across all industrial sectors. 
Our first observation is that the robustness scores for a specific ability vary significantly across LLMs. 
For example, the highest robustness score for the ``Magnitude'' ability is 0.9167, achieved by the ``AN_Model 4'' model, while the ``Tiangong'' model only achieves a score of 0.3333.
Additionally, when we focus on a specific LLM, we observe that the robustness scores can vary across abilities, unlike the performance gap between industrial sectors. For instance, the ``AN_Model 4'' model can achieve a robustness score of 0.9686 for the ``Synonyms'' ability, but only 0.4552 for the ``Logic'' ability. Similar trends are observed for other LLMs, where logic ability robustness is generally lower than other abilities, around 0.6. 
This highlights the varying difficulties of variants generated by MT relations. The synonyms capability is basic, requiring only the understanding of question word synonyms, while the Logic capability is more complex, demanding comprehension of industrial questions along with diverse expressions of equivalent or opposite meanings.

\parh{Ability Gap between Local and Global LLMs.}~We further delve into the robustness scores of specific abilities between local and global LLMs to gain a more nuanced understanding of their robustness.
Surprisingly, our findings reveal that the overall robustness advantage of GPT models over local LLMs does not necessarily extend to specific abilities. For example, for the ``Synonyms'' and ``Magnitude'' abilities, the highest scores are attained by the ``AN_Model 4'' model rather than the GPT series. Additionally, other local models exhibit competitive robustness on these abilities, despite lower overall robustness and accuracy.
We believe this is due to the fact that local vendors prioritize serving Chinese users and may not prioritize advanced abilities such as complex reasoning. Therefore, it is reasonable to assume that local LLMs are better suited for abilities that require an understanding of industry-specific terminology.

\begin{figure}[!htpb]
    \centering
    \includegraphics[width=0.49\textwidth]{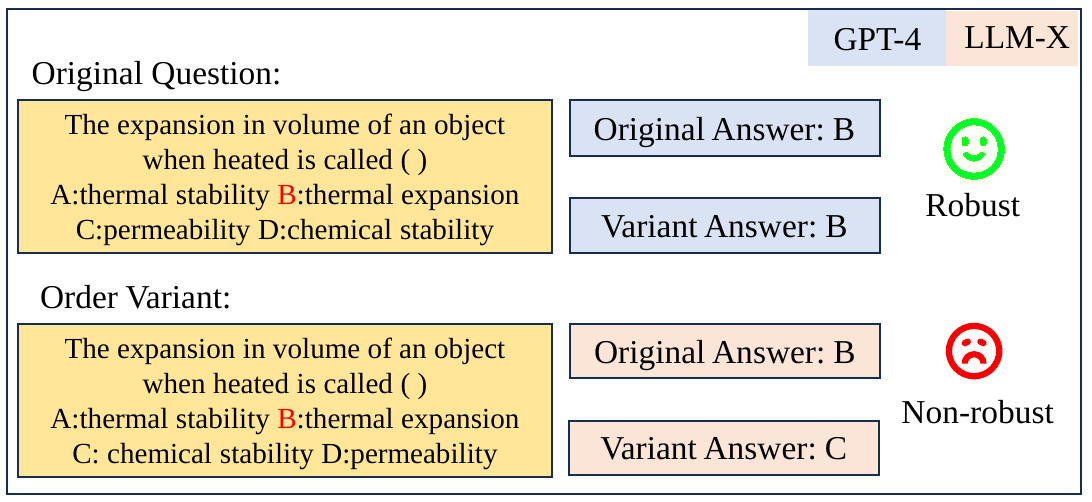} 
    \vspace{-10pt}
    \caption{The accuracy of LLMs in the Chinese industrial scenarios.}
    \vspace{-10pt}
    \label{fig:example-ability}
\end{figure}

However, there are also abilities where GPT models clearly outperform local LLMs. For example, for ``Order'' and ``Logic'' abilities, GPT robustness scores substantially exceed local models, with average relative improvements of 36.78\% and 40.03\% respectively.
To better illustrate this, we provide an example in~\F~\ref{fig:example-ability}, where the GPT-4 model achieves a robust result by correctly answering a question and its variant, while a local LLM fails to do so. The variant in this example is generated by the ``Order'' relation, which simply exchanges the content of option C and option D. Although not perfect, we find that GPT-4 correctly answers most of the variants generated by the ``Order'' relation, which is in line with previous work~\cite{zheng2023large}.
We attribute this to the fact that GPT-4 model can achieve a robust result because it has a better understanding of the question, and can better capture the relationship between the question and the options. In contrast, the local LLMs are more likely to be affected by the change of the order of the options, and thus fail to answer the question correctly.
It is worth noting that there are also abilities in which both local and global LLMs achieve competitive robustness scores. For instance, in the ``Security'' and ``General'' abilities, local LLMs can also achieve competitive scores, suggesting that these abilities are less complex and require an understanding of terminology to aid variant answering.

\finding{4}{The scores for a particular ability differ greatly among LLMs, and vice versa, the robustness scores of LLMs vary significantly across abilities. Local LLMs show competitive or higher performance than GPT series on abilities that require an understanding of special terms in industrial scenarios, while GPT series have a clear advantage on abilities related to logic and reasoning.}

\section{Related Work}
\label{sec:related}

\parh{LLM Benchmarks.}~LLM benchmarks play a crucial role as standardized tests, enabling comparison of diverse models across multiple tasks. With the field's ongoing progress, numerous benchmarks have arisen~\cite{chatbotarena,Huang2023WhoIC,Yuan2023GPT4IT,zhong2023agieval,leaderboard,zheng2023judging,Wang2024TheEI}, which can be classified into two primary categories: general language benchmarks and task-specific benchmarks. General language benchmarks, such as Chatbot Arena~\cite{chatbotarena} and MT-Bench~\cite{zheng2023judging}, facilitate the assessment and improvement of chatbot models and LLMs within various contexts. Chatbot Arena offers a competitive environment to evaluate and compare heterogeneous chatbot models based on user interaction and voting. Conversely, MT-Bench examines LLMs in multi-turn dialogues utilizing comprehensive questions designed for conversational handling. MMLU~\cite{hendrycks2020measuring} is another benchmark offering an extensive set of tests for gauging text models in multi-task settings, while AlpacaEval~\cite{alpacaeval} centers on evaluating LLMs across diverse NLP tasks. AGIEval~\cite{zhong2023agieval} acts as a specialized evaluation framework dedicated to assessing the performance of foundation models in human-centric standardized exams. Moreover, OpenLLM~\cite{leaderboard} operates as an evaluation benchmark by presenting an open competition platform that measures and compares distinct LLM models' performance across myriad tasks, fostering development and competition in LLM research.

Besides the general language benchmarks, task-specific benchmarks~\cite{singhal2022large,ji2023benchmarking,sawada2023arb,huang2023trustgpt,huang2023ceval} have been developed to assess LLMs within distinct downstream applications. For example, MultiMedQA~\cite{singhal2022large} concentrates on medical question-answering, while ARB~\cite{sawada2023arb} appraises the ability of LLMs to perform advanced reasoning tasks spanning multiple domains. TRUSTGPT~\cite{huang2023trustgpt} explicitly deals with ethical concerns, and EmotionBench~\cite{huang2023emotionally} evaluates the capacity of LLMs to mimic human emotional responses. Additionally, C-Eval~\cite{huang2023ceval} measures the advanced knowledge and reasoning proficiencies of foundational models in Chinese, whereas M3Exam~\cite{zhang2023m3exam} examines the general abilities of LLMs across diverse contexts utilizing multiple languages and modalities.

Notably, the benchmark proposed in this paper differs from existing benchmarks by being the first to specifically focus on evaluating the accuracy and robustness of LLMs for industrial sector applications. While prior benchmarks have evaluated aspects such as reasoning and ethics, our study is innovative in determining the suitability of LLMs for real-world applications in industry-specific verticals. The benchmark delivers a standardized evaluation of the accuracy and reliability with which LLMs can support tasks in essential industrial domains.

\parh{Automatic Evaluation of AI Outputs.}~In light of the rapid advancement of LLMs, it's crucial to establish dependable and sturdy evaluation techniques for measuring AI-generated outputs, particularly for open-ended inquiries. Traditional automated metrics like BLEU~\cite{papineni2002bleu} and ROUGE~\cite{lin2004rouge} are widely employed; however, they exhibit shortcomings in gauging semantic meaning, reasoning, and logical coherence~\cite{zhang2019bertscore,wang2022enriching,gu2021cradle,li2022unleashing,ma2023oops,peng2022static}. Lately, neural representation methods have been investigated to create more reliable semantic similarity measures, but imperfections persist. As an alternative automated evaluation method, LLMs have recently gained popularity~\cite{ganguli2023challenges,zheng2023judging,wang2023pandalm,chan2023chateval}. LLM-EVAL~\cite{lin-chen-2023-llm} offers a comprehensive framework for evaluating conversational quality in various aspects. PandaLM~\cite{wang2023pandalm} compiles an assortment of human annotations to train a model capable of predicting unbiased assessments in a replicable manner. Furthermore, several studies~\cite{chan2023chateval,zheng2023judging} have adapted the Elo rating system that was initially designed for chess to assess LLM performance~\cite{elo1978rating}. By considering LLMs as players in pairwise matchups simulated using LLM-based evaluators, the emerging Elo ratings enable a solid comparison of models across numerous virtual contests. In~\S~\ref{subsec:question-format}, we discussed comparing the outputs of LLMs with ground truth to evaluate their performance, thus avoiding the limitations of traditional automated measures and yielding more trustworthy results.

\parh{Metamorphic Testing.}~Metamorphic testing (MT)~\cite{chen1998metamorphic} refers to a testing approach that employs metamorphic relations to identify potential defects in software applications. Owing to its effectiveness in bug detection~\cite{chen2018mtr}, MT has been widely applied in various fields, such as autonomous vehicles~\cite{zhang2018deeproad}, visual question answering (VQA)~\cite{yuan2021perception}, text moderation software~\cite{Wang2023MTTMMT}, multimodel reasoning~\cite{li2023vrptest} and object identification~\cite{wang2020metamorphic}. In recent times, MT has been used for LLM-based systems to pinpoint potential problems.
For instance,~\citet{li2022cctest} proposed CCTEST, a testing framework for code completion that checks for errors in systems such as GitHub Copilot~\cite{copilot}. The framework generates different versions of code fragments and examines inconsistencies in their syntax and semantics. 
Additionally,~\citet{ma2021mt} developed an MT approach for Natural Language Interface to Database (NLIDB) systems, and discovered thousands of defects.
In this paper, we design eight industry-specific metamorphic relations for LLMs and apply them to evaluate the robustness of LLMs in industrial applications.

\parh{LLM Robustness.}~Investigating the stability of systems when confronted with unexpected inputs is the focus of robustness research. Two primary aspects of robustness are out-of-distribution (OOD) robustness and adversarial robustness~\cite{chang2023survey}. OOD robustness examines a model's performance on inputs deviating from the training distribution, whereas adversarial robustness gauges its resilience against purposely introduced minor perturbations aimed at misleading the model~\cite{nie2019adversarial}. There are numerous benchmarks and approaches designed for assessing robustness, such as AdvGLUE~\cite{wang2021adversarial}, ANLI~\cite{nie2019adversarial}, and DDXPlus~\cite{fansi2022ddxplus}. For instance, \citet{wang2023robustness} appraise the adversarial and OOD robustness of ChatGPT and other extensive language models using these benchmarks. In terms of methodology, \citet{yang2022glue} choose to incorporate OOD examples into the GLUE benchmark~\cite{wang2018glue}. More precisely, \citet{li2023survey} deliver a comprehensive examination of OOD evaluation, consolidating adversarial robustness, domain generalization, data biases, and proposing prospective research areas.

Adversarial robustness has emerged as another major emphasis in robustness studies, particularly as a result of recent jailbreak attacks on sophisticated LLMs like ChatGPT that demonstrate substantial risks for societal impact~\cite{liu2023jailbreaking}.
In order to methodically assess adversarial robustness,~\citet{zhu2023promptbench} present a consolidated benchmark named PromptBench dedicated to comprehensive adversarial text attacks. Their findings indicate that current LLMs remain vulnerable to adversarial prompts, highlighting the necessity for improving these models' resilience against such inputs. 
With respect to novel adversarial datasets,~\citet{wang2023decodingtrust} unveil the AdvGLUE++ dataset for evaluating adversarial robustness and develop an innovative evaluation protocol to critically examine machine ethics via jailbreaking system prompts.
Identifying robustness issues in this inquiry is relatively uncomplicated compared to other areas like reasoning and ethics, as the ground truth answer can be compared to the produced content, facilitating accurate judgment. Nevertheless, the consequences of robustness problems are graver, considering that this research centers on the industrial generation process, where both exceptional quality and robustness are crucial. As a result, the importance of robustness is paramount in guaranteeing the dependability and efficacy of the industrial generation process.

\section{Discussion}
\label{sec:discussion}

\parh{Threats to Validity.}~The validity of the results presented in this research may be subject to certain limitations. 
First, while we have made efforts to evaluate a diverse set of popular LLMs across major industrial sectors, the rapidly evolving landscape of LLMs implies that newly released models or updated versions of existing ones might be excluded from our analysis. Additionally, the closed-source nature of many commercial LLMs makes it challenging to promptly identify their latest iterations. However, we believe our proposed benchmark and framework establish a solid basis for assessing LLM capabilities across industrial scenarios. 
The second concern we need to address is the extent of coverage of the sectors that are under evaluation.
While we carefully choose representative industrial sectors, it is important to acknowledge that each sector requires specialized expertise and knowledge that may not be fully captured in our evaluation. Additionally, it is possible that our sample may not fully encompass all sector-specific complexities across the board. Despite attempts to pick representative sectors and formulate illustrative relations, the evaluation methodology merits further enhancement to encompass diverse domains.
Lastly, the recommended configurations and parameters for experiments could potentially impact outcomes. To ensure replicable and comparable evaluations, we fix the ``temperature'' at 0 for cloud-based LLMs and disable sampling for locally deployed models. However, as advocated by most LLM providers, such settings may not optimize real-world effectiveness. Hence, observations derived from our regulated environment may deviate from LLMs implemented under default settings calibrated by vendors. Moving forward, it would be prudent to calibrate configurations to vendor guidelines for holistic appraisals in applied contexts.

\parh{Framework Extension.}~Our MT-based framework aims to assess LLMs within Chinese industrial applications, and the framework can be readily extended to other languages and fields.
For instance, metamorphic relations for English LLMs can be established to evaluate their efficacy in English industrial applications. Importantly, in cases involving low-resource languages with insufficient assessment data available, test scenarios could leverage English data translated into the target language.
In addition to linguistic expansion, the methodology can be adapted to other sectors like medicine, law, and finance by creating domain-specific seed questions and relations.
In conclusion, the framework's adaptability regarding languages and fields renders it appropriate for examining LLMs across a broad range of contexts and applications. Through proper alterations to test datasets and relations, the methodology can adapt to emerging problem domains while maintaining stringent evaluation criteria.

\parh{Future Work.}~This study primarily concentrates on the precision and resilience of LLMs in Chinese industrial scenarios. Nevertheless, additional critical factors warrant investigation, including LLMs' efficiency, privacy, and security. Assessing the computational and inference performance of LLMs for industrial tasks allows enterprises to identify ideal models that meet their latency, throughput, and cost objectives. Establishing efficiency benchmarks tailored to industry-specific scenarios can facilitate adoption by clarifying the trade-offs between accuracy and processing speed.
Privacy assurance is crucial when handling personal or confidential information prevalent in certain sectors. Investigating the privacy risks associated with LLMs and exploring innovative privacy-enhancing methods such as watermarking~\cite{li2023protecting,szyller2021dawn,cong2022sslguard}, federated learning~\cite{pang2022adi,shao2023survey,sun2023semi}, differential privacy~\cite{ma2022noleaks}, and homomorphic encryption~\cite{gentry2009fully} for industrial applications is of paramount importance. 
Furthermore, the security of LLMs is another area of concern, especially when their output code snippets~\cite{li2024evaluating,DBLP:conf/kbse/WenWGWLG23} or decisions~\cite{tamkin2023evaluating} are deployed in critical industrial systems. 
Overall, this research serves as a foundational step toward gaining comprehensive insight into LLMs within industrial contexts. We aim to encourage further initiatives that focus on more than just accuracy by employing multidimensional benchmarks and strategies that reveal trade-offs, biases, and potential hazards. Strengthening collaboration between industry and academia is essential for developing dependable and resilient LLM solutions that address real-world demands.

\section{Conclusion}
\label{sec:conclusion}

In this paper, we present a comprehensive empirical study to assess the accuracy and robustness of LLMs in industrial scenarios. 
We collect the first benchmark of industry-specific problems in Chinese across eight different industrial sectors, and propose a metamorphic testing framework with industry-specific metamorphic relations to assess robustness.
Massive experiments are conducted to evaluate 10 LLMs from 9 different vendors, and compare the local LLMs with global LLMs in terms of multiple abilities and sectors.
Our findings provide guidance for developing LLMs that better serve non-English (Chinese) users in industrial applications. These findings also assist platform engineers and enterprises in improving Chinese LLMs for manufacturing.

\balance
\bibliographystyle{ACM-Reference-Format}
\bibliography{bib/ref,bib/analysis,bib/decompiler,bib/testing-cv,bib/cv,bib/reference,bib/llm,bib/zj,bib/sw,bib/wx}

\end{document}